\documentclass[twoside,11pt]{article}

\usepackage{blindtext}

\usepackage{jmlr2e}

\usepackage{lastpage}

\firstpageno{1}

\usepackage{tabularx}
\usepackage{graphicx}
\usepackage{amsmath}
\usepackage{caption}
\usepackage{subcaption}

\begin{document}

\title{A Comparison of Lightweight Deep Learning Models for Particulate-Matter Nowcasting in the Indian Subcontinent \& Surrounding Regions}

\author{\name Ansh Kushwaha \email ansh.kushwaha@flame.edu.in \\
       \addr Centre for Interdisciplinary Artificial Intelligence (CAI)\\
       FLAME University, Pune, Maharashtra 412115, India
       \AND
       \name Kaushik Gopalan \email kaushik.gopalan@flame.edu.in \\
       \addr Centre for Interdisciplinary Artificial Intelligence (CAI) \& School of Computing and Data Sciences\\
       FLAME University, Pune, Maharashtra 412115, India}



\maketitle

\begin{abstract}
This paper is a submission for the Weather4Cast~2025 complementary Pollution Task and presents an efficient framework for 6-hour lead-time nowcasting of PM$_1$, PM$_{2.5}$, and PM$_{10}$ across the Indian subcontinent and surrounding regions. The proposed approach leverages analysis fields from the Copernicus Atmosphere Monitoring Service (CAMS) Global Atmospheric Composition Forecasts at 0.4$^{\circ}$ resolution. A 256$\times$256 spatial region, covering 28.4$^{\circ}$S--73.6$^{\circ}$N and 32.0$^{\circ}$E--134.0$^{\circ}$E, is used as the model input, while predictions are generated for the central 128$\times$128 area spanning 2.8$^{\circ}$S--48.0$^{\circ}$N and 57.6$^{\circ}$E--108.4$^{\circ}$E, ensuring an India-centric forecast domain with sufficient synoptic-scale context. Models are trained on CAMS analyses from 2021--2023 using a shuffled 90/10 split and independently evaluated on 2024 data. Three lightweight parameter-specific architectures are developed to improve accuracy, minimize systematic bias, and enable rapid inference. Evaluation using RMSE, MAE, Bias, and SSIM demonstrates substantial performance gains over the Aurora foundation model [\cite{bodnar2025aurora}], underscoring the effectiveness of compact \& specialized deep learning models for short-range forecasts on limited spatial domains.
\end{abstract}


\footnotetext{Code and Trained Models: \url{https://github.com/flame-cai/w4c25_pollution}}

\section{Introduction}

Accurate short-term forecasting of particulate matter (PM$_{1}$, PM$_{2.5}$, PM$_{10}$) is essential to protect public health, ecosystems and regional climate. Although recent Earth-system foundation models such as Aurora \cite{bodnar2025aurora} have advanced global atmospheric prediction through physics-informed learning, their computational scale and general-purpose design limit their practicality for regional high-frequency applications. Efficient and localized models are therefore required to address the rapidly varying air-quality conditions over the Indian subcontinent. This paper is a submission for the Weather4Cast~2025 complementary Pollution Task and focuses on developing compact architectures capable of delivering accurate and operationally feasible 6-hour particulate-matter nowcasts.

Recent research has increasingly focused on deep learning approaches for regional air-quality prediction, with CNN–RNN and BiLSTM architectures demonstrating strong capability in modeling complex spatiotemporal pollutant behaviour (See \cite{utku2025advancing, lei2025shap}). Region-specific studies across China (See \cite{liu2025deep}), Thailand (See \cite{damkliang2025deep}), and the UAE (See \cite{abuouelezz2025exploring}) highlight the advantages of localized learning when forecasting fine-scale particulate dynamics. In the Indian subcontinent, stacked and ensemble deep models (See \cite{patel2025stacked, patel2025systematic}) have shown improvements in capturing regional variability arising from diverse meteorological and emission conditions. Additional deep-learning advancements include explainable PSO-CNN-BiLSTM networks (See \cite{lei2025shap}), SARIMA–BiLSTM hybrids (See \cite{necula2025advanced}), physics-constrained PCDCNet (See \cite{wang2025pcdcnet}), graph-based E-STGCN models (See \cite{panja2024stgcn}), and robust ensemble frameworks (See \cite{petric2024ensemble}). 

Despite these developments, efficient 6-hour particulate-matter nowcasting using CAMS analysis fields remains largely unexplored, reinforcing the need for lightweight, region-optimized architectures designed for high-frequency operational forecasting.

\section{Methodology}


This study employs analysis data from the Copernicus Atmosphere Monitoring Service (CAMS) Global Atmospheric Composition Forecasts, generated operationally by European Centre for Medium-Range Weather Forecasts (ECMWF) using the Integrated Forecasting System (IFS) with four-dimensional variational (4D-VAR) data assimilation. The analyses are provided for four timestamps a day (00, 06, 12, and 18~UTC) at a uniform horizontal resolution of 0.4$^{\circ}$. From this dataset, ten surface-level variables relevant to near-surface meteorology and particulate-matter behavior were selected: 2\,m temperature, 2\,m dewpoint temperature, 10\,m zonal and meridional wind components, mean sea-level pressure, land--sea mask, surface geopotential, and concentrations of PM$_{1}$, PM$_{2.5}$, and PM$_{10}$. Data covering the period 2021--2024 were retrieved in NetCDF format, with each file corresponding to a single day. Only this subset of CAMS was used for training, validation, and testing.


For spatial domain selection, a 256$\times$256 grid-cell region was extracted from the global field, spanning 28.4$^{\circ}$S--73.6$^{\circ}$N and 32.0$^{\circ}$E--134.0$^{\circ}$E. Model predictions were subsequently restricted to the central 128$\times$128 domain, covering 2.8$^{\circ}$S--48.0$^{\circ}$N and 57.6$^{\circ}$E--108.4$^{\circ}$E, defining an India-centric output region with sufficient synoptic-scale context. All variables were normalized by dividing each field by its global absolute maximum value, constraining inputs to the range [$-1$, 1] while preserving directional sign information for wind components. The normalized variables were stacked into ten-channel tensors representing individual analysis timestamps. Data from 2021--2023 were randomly shuffled and split into 90\% for training and 10\% for validation, while the entire year 2024 was held out as an independent test set to assess seasonal and interannual generalization.


Each model predicts a single-parameter field (PM$_1$, PM$_{2.5}$, or PM$_{10}$) at a 6-hour lead time using the ten-channel input from a single CAMS analysis. Separate models were trained for each parameter to ensure independent optimization. ConvGRU and ConvLSTM architectures used a composite loss emphasizing accuracy and spatial structure:
\[
\mathcal{L} = 0.75\times\frac{\mathrm{RMSE}}{\mathrm{mean}} + 0.25\times(1 - \mathrm{SSIM}),
\]
while U-Net employed a Huber-based variant for improved gradient stability:
\[
\mathcal{L} = 0.7\times\frac{\mathrm{Huber}}{\mathrm{mean}} + 0.3\times(1 - \mathrm{SSIM}).
\]
All models were trained with a batch size of 32 and without data augmentation. ConvGRU/ConvLSTM networks used the Adam optimizer (learning rate\,=\,$10^{-3}$) with a ReduceLROnPlateau scheduler (factor\,=\,0.5, patience\,=\,3). U-Net training used AdamW (weight\,=\,$10^{-4}$) with cosine-annealing scheduling and two-step gradient accumulation.


\begin{table}[h!]
\caption{Model Configurations Summary}
\label{tab:models}
\begin{tabularx}{\textwidth}{| >{\hsize=.5\hsize\linewidth=\hsize}X | c | >{\hsize=1.5\hsize\linewidth=\hsize}X |}
\hline
\textbf{Model \textit{(Type)}} & \textbf{Parameters} & \textbf{Description} \\
\hline
Aurora \newline \textit{(Foundation)} 
& 1.3B 
& Global physics-informed Earth-system foundation model using 3D Swin Transformer and Perceiver-based encoders; serves as baseline \newline [\cite{bodnar2025aurora}]. \\
\hline
ConvGRU$_{1}$ \newline \textit{(Baseline)} 
& 1.12M 
& In: (\(\mathcal{B}\), \(\mathcal{T}\), \(\mathcal{C}\), 256, 256); H = [64, 128, 64]; \newline $k$=3; Dropout = [0.2, 0.3, 0.2]; \newline Out: (\(\mathcal{B}\), 1, 128, 128) via 1$\times$1 Conv + Tanh. \\
\hline
ConvGRU$_{2}$ \newline \textit{(Baseline)} 
& 1.47M 
& In: (\(\mathcal{B}\), \(\mathcal{T}\), \(\mathcal{C}\), 256, 256); H = [128, 64, 128]; \newline $k$=3; Dropout = [0.3, 0.2, 0.3]; \newline Out: (\(\mathcal{B}\), 1, 128, 128) via 1$\times$1 Conv + Tanh. \\
\hline
ConvLSTM$_{1}$ \newline \textit{(Baseline)} 
& 1.50M 
& In: (\(\mathcal{B}\), \(\mathcal{T}\), \(\mathcal{C}\), 256, 256); H = [64, 128, 64]; \newline $k$=3; Dropout = [0.2, 0.3, 0.2]; \newline Out: (\(\mathcal{B}\), 1, 128, 128) via 1$\times$1 Conv + Tanh. \\
\hline
ConvLSTM$_{2}$ \newline \textit{(Baseline)} 
& 1.96M 
& In: (\(\mathcal{B}\), \(\mathcal{T}\), \(\mathcal{C}\), 256, 256); H = [128, 64, 128]; \newline $k$=3; Dropout = [0.3, 0.2, 0.3]; \newline Out: (\(\mathcal{B}\), 1, 128, 128) via 1$\times$1 Conv + Tanh. \\
\hline
U-Net \newline \textit{(Proposed)} 
& 4.33M 
& In: (\(\mathcal{B}\), \(\mathcal{T}\), \(\mathcal{C}\), 256, 256) $\rightarrow$ (\(\mathcal{B}\), \(\mathcal{T}\)$\times$\(\mathcal{C}\), 256, 256); \newline Encoder: [16, 32, 64, 128, 256]; \newline Decoder: bilinear upsampling + skip connections; \newline Out: (\(\mathcal{B}\), 1, 128, 128) via 1$\times$1 Conv + Tanh. \\
\hline
\end{tabularx}
\end{table}

Five lightweight models were evaluated against the 1.3B-parameter Aurora foundation model [\cite{bodnar2025aurora}]. Although Aurora delivers state-of-the-art global atmospheric forecasts, we hypothesize that comparable performance can be attained by restricting spatial domains and for short lead times using much smaller architectures. Consequently, the proposed ConvGRU, ConvLSTM and U-Net models—each with only 1--4,M parameters—are optimized for fast, resource-efficient inference and specifically tailored for 6-hour particulate-matter nowcasting over the Indian region, offering a practical and computationally affordable alternative to large foundation models.

The ConvGRU and ConvLSTM variants employ three stacked convolutional recurrent layers (kernel size\,=\,3) with dropout regularization and a final 1$\times$1 convolution to produce the 128$\times$128 forecast domain. The U-Net follows an encoder–decoder structure with bilinear upsampling and skip connections, enabling effective multi-scale feature reconstruction from the 256$\times$256 input patch. A complete summary of each model’s configuration is provided in Table~\ref{tab:models}. All model implementations, training pipelines, preprocessing scripts, and trained checkpoints are openly available at: \url{https://github.com/flame-cai/w4c25_pollution}.

\section{Results}

Tables~\ref{tab:pm1}--\ref{tab:pm10} present model performance for PM$_1$, PM$_{2.5}$, and PM$_{10}$, with RMSE, MAE, and Bias reported in \(\mu\mathrm{g}\,\mathrm{m}^{-3}\). The proposed models consistently surpass Aurora, achieving lower errors, higher SSIM, and reduced bias. Figures~\ref{fig:pm_metrics_sidebyside} and~\ref{fig:pm_bias} further confirm robust seasonal stability and improved spatial fidelity.

\subsection{PM$_{1}$ Forecasting Performance}

\begin{table}[h!]
\centering
\caption{PM$_{1}$ Evaluation Metrics at 12:00 UTC Input and Diurnal Average}
\label{tab:pm1}
\begin{tabular}{l | cccc | cccc}
\hline
 & \multicolumn{4}{c|}{12:00 UTC Input} & \multicolumn{4}{c}{Diurnal Average} \\
Model & RMSE & MAE & Bias & SSIM & RMSE & MAE & Bias & SSIM \\
\hline
Aurora & 9.04 & 7.28 & 5.41 & 0.48 & - & - & - & - \\
ConvGRU$_{1}$ & 5.58 & 2.92 & 0.39 & 0.80 & 4.83 & 2.74 & 0.86 & 0.79 \\
ConvGRU$_{2}$ & 5.44 & 2.71 & 0.14 & 0.82 & 4.75 & 2.58 & 0.64 & 0.81 \\
ConvLSTM$_{1}$  & 5.67 & 2.75 & -0.32 & 0.81 & 4.83 & 2.56 & 0.13 & 0.80 \\
ConvLSTM$_{2}$ & 5.44 & 2.66 & 0.05 & 0.83 & 4.73 & 2.53 & 0.54 & 0.81 \\
U-Net & 4.81 & 2.42 & 0.11 & 0.83 & 4.53 & 2.41 & 0.06 & 0.81 \\
\hline
\end{tabular}
\end{table}

Table~\ref{tab:pm1} shows that Aurora produces an RMSE of 9.04 \(\mu\mathrm{g}\,\mathrm{m}^{-3}\) and SSIM of 0.48, while the proposed models reduce RMSE by 35--47\%, MAE by 60--67\%, and absolute bias by more than 98\% at 12:00~UTC. ConvLSTM$_2$ achieves an RMSE of 5.44 \(\mu\mathrm{g}\,\mathrm{m}^{-3}\) (39.9\% reduction) and SSIM of 0.83 (72.9\% improvement). U-Net yields the strongest overall performance with RMSE 4.81 \(\mu\mathrm{g}\,\mathrm{m}^{-3}\) and SSIM 0.83. Diurnal averages demonstrate similar patterns, with U-Net maintaining the lowest RMSE (4.53 \(\mu\mathrm{g}\,\mathrm{m}^{-3}\)) and a near-zero bias (0.06), confirming consistently high predictive accuracy across all times of day.

\subsection{PM$_{2.5}$ Forecasting Performance}

\begin{table}[h!]
\centering
\caption{PM$_{2.5}$ Evaluation Metrics at 12:00 UTC Input and Diurnal Average}
\label{tab:pm2p5}
\begin{tabular}{l | cccc | cccc}
\hline
 & \multicolumn{4}{c|}{12:00 UTC Input} & \multicolumn{4}{c}{Diurnal Average} \\
Model & RMSE & MAE & Bias & SSIM & RMSE & MAE & Bias & SSIM \\
\hline
Aurora & 12.26 & 6.52 & 0.90 & 0.75 & - & - & - & - \\
ConvGRU$_{1}$ & 10.27 & 4.72 & 0.63 & 0.85 & 11.37 & 4.70 & 1.23 & 0.83 \\
ConvGRU$_{2}$ & 10.41 & 4.32 & -0.86 & 0.87 & 11.20 & 4.33 & -0.34 & 0.84 \\
ConvLSTM$_{1}$  & 10.39 & 4.55 & 0.14 & 0.86 & 11.25 & 4.52 & 0.68 & 0.83 \\
ConvLSTM$_{2}$ & 10.41 & 4.66 & 0.15 & 0.86 & 11.69 & 4.67 & 0.61 & 0.82 \\
U-Net & 9.88 & 4.22 & -0.46 & 0.87 & 10.38 & 4.38 & -0.14 & 0.83 \\
\hline
\end{tabular}
\end{table}

As shown in Table~\ref{tab:pm2p5}, Aurora’s RMSE of 12.26 \(\mu\mathrm{g}\,\mathrm{m}^{-3}\) and SSIM of 0.75 are surpassed by all proposed models, which deliver 15--20\% RMSE reduction, 28--35\% MAE reduction, and up to 95\% lower absolute bias. ConvGRU$_2$ reaches the highest SSIM (0.87), while U-Net attains the lowest RMSE (9.88 \(\mu\mathrm{g}\,\mathrm{m}^{-3}\)). Diurnal averages further confirm that U-Net remains stable with an RMSE of 10.38 \(\mu\mathrm{g}\,\mathrm{m}^{-3}\) and very low bias (0.14), showing strong temporal generalization.

\subsection{PM$_{10}$ Forecasting Performance}

\begin{table}[h!]
\centering
\caption{PM$_{10}$ Evaluation Metrics at 12:00 UTC Input and Diurnal Average}
\label{tab:pm10}
\begin{tabular}{l | cccc | cccc}
\hline
 & \multicolumn{4}{c|}{12:00 UTC Input} & \multicolumn{4}{c}{Diurnal Average} \\
Model & RMSE & MAE & Bias & SSIM & RMSE & MAE & Bias & SSIM \\
\hline
Aurora & 67.47 & 29.75 & 18.08 & 0.76 & - & - & - & - \\
ConvGRU$_{1}$ & 53.32 & 15.13 & 2.95 & 0.91 & 52.91 & 14.55 & 1.61 & 0.90 \\
ConvGRU$_{2}$ & 54.45 & 16.54 & -8.00 & 0.86 & 53.68 & 16.96 & -9.34 & 0.83 \\
ConvLSTM$_{1}$  & 53.39 & 13.83 & 0.22 & 0.92 & 52.39 & 13.67 & -1.17 & 0.91 \\
ConvLSTM$_{2}$ & 54.27 & 14.37 & 1.96 & 0.92 & 53.58 & 14.00 & 0.47 & 0.91 \\
U-Net & 53.10 & 14.04 & 0.99 & 0.92 & 49.58 & 13.67 & -0.56 & 0.91 \\
\hline
\end{tabular}
\end{table}

Table~\ref{tab:pm10} highlights Aurora’s comparatively weak PM$_{10}$ performance (RMSE 67.47 \(\mu\mathrm{g}\,\mathrm{m}^{-3}\), SSIM 0.76). All proposed models achieve 20--26\% lower RMSE, 50--54\% lower MAE, and 98--99\% lower absolute bias. ConvLSTM$_1$ and U-Net achieve SSIM values of 0.92, marking a 21\% improvement in structural fidelity. In diurnal averages, U-Net again exhibits the best RMSE (49.58 \(\mu\mathrm{g}\,\mathrm{m}^{-3}\)) and very low bias (0.56), demonstrating superior daily stability.

\subsection{Seasonal Performance Variability}

\begin{figure}[h!]
    \centering
    \begin{subfigure}{0.49\textwidth}
        \includegraphics[width=\textwidth]{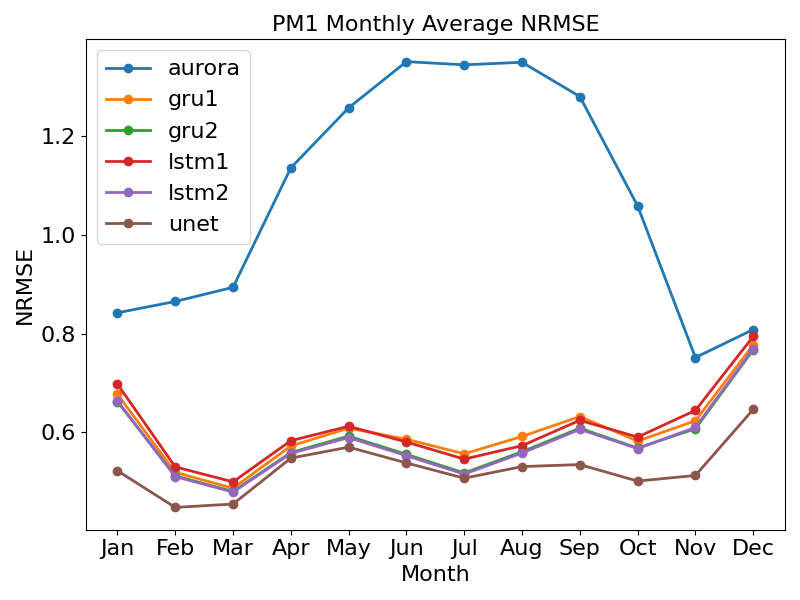}
        \caption{PM$_{1}$ NRMSE}
    \end{subfigure}
    \hfill
    \begin{subfigure}{0.49\textwidth}
        \includegraphics[width=\textwidth]{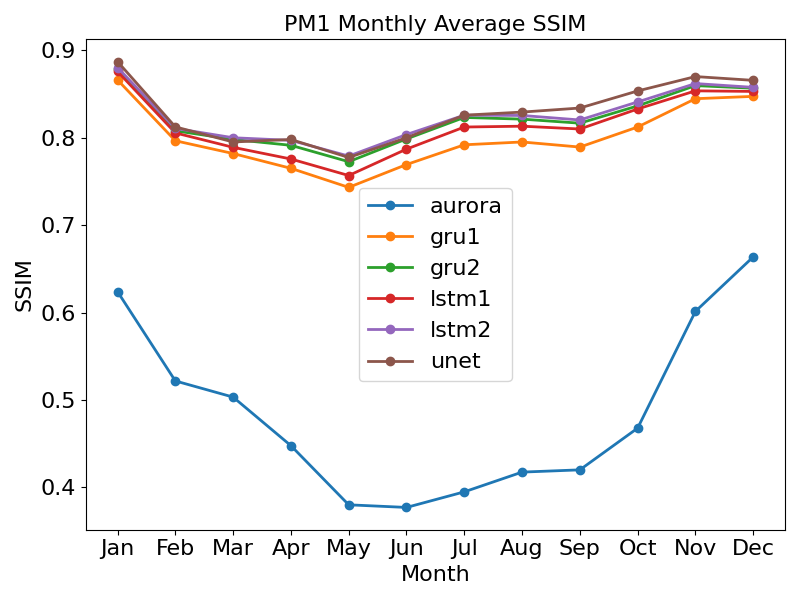}
        \caption{PM$_{1}$ SSIM}
    \end{subfigure}
    \begin{subfigure}{0.49\textwidth}
        \includegraphics[width=\textwidth]{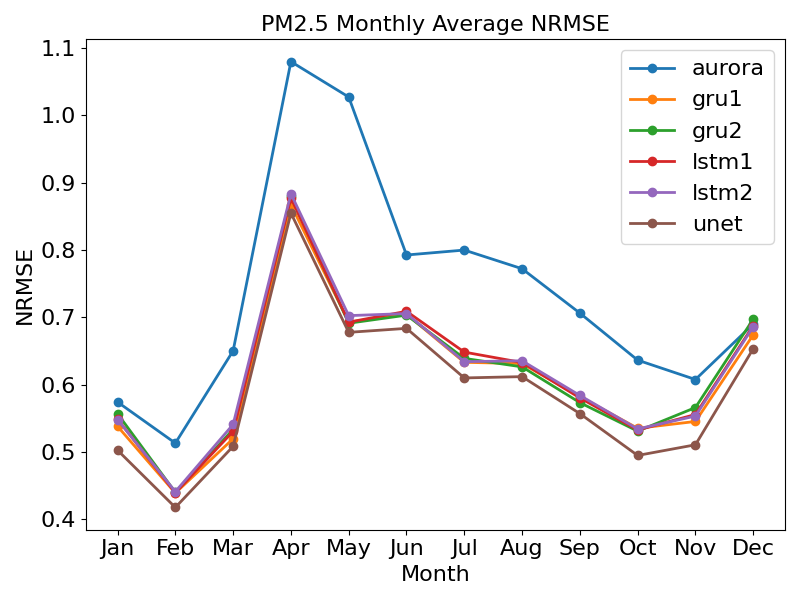}
        \caption{PM$_{2.5}$ NRMSE}
    \end{subfigure}
    \hfill
    \begin{subfigure}{0.49\textwidth}
        \includegraphics[width=\textwidth]{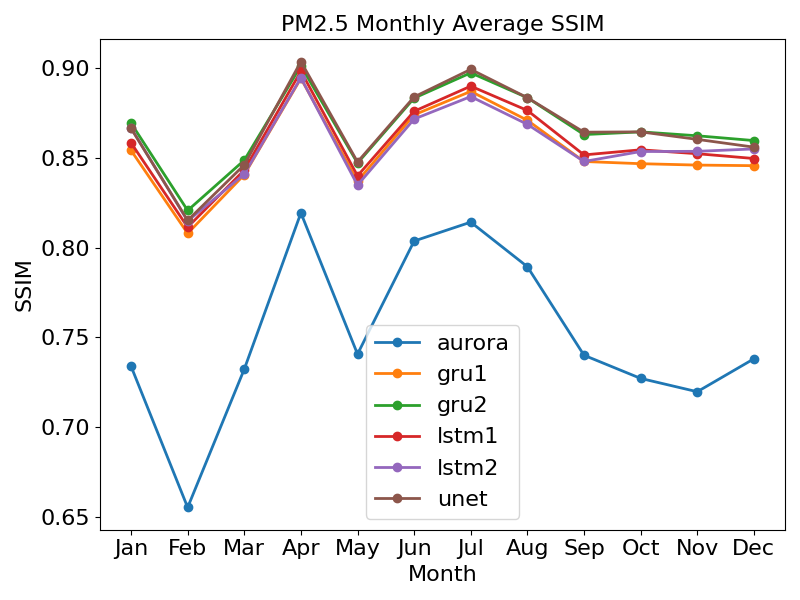}
        \caption{PM$_{2.5}$ SSIM}
    \end{subfigure}
    \begin{subfigure}{0.49\textwidth}
        \includegraphics[width=\textwidth]{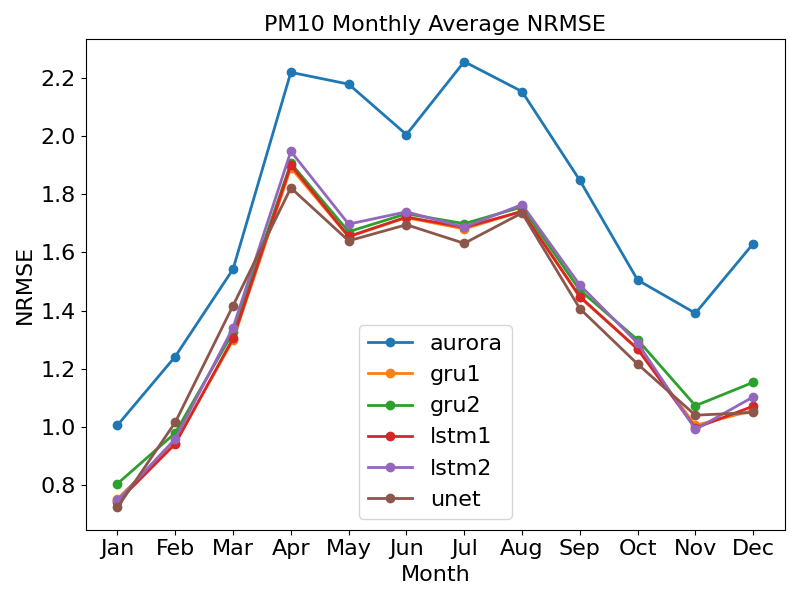}
        \caption{PM$_{10}$ NRMSE}
    \end{subfigure}
    \hfill
    \begin{subfigure}{0.49\textwidth}
        \includegraphics[width=\textwidth]{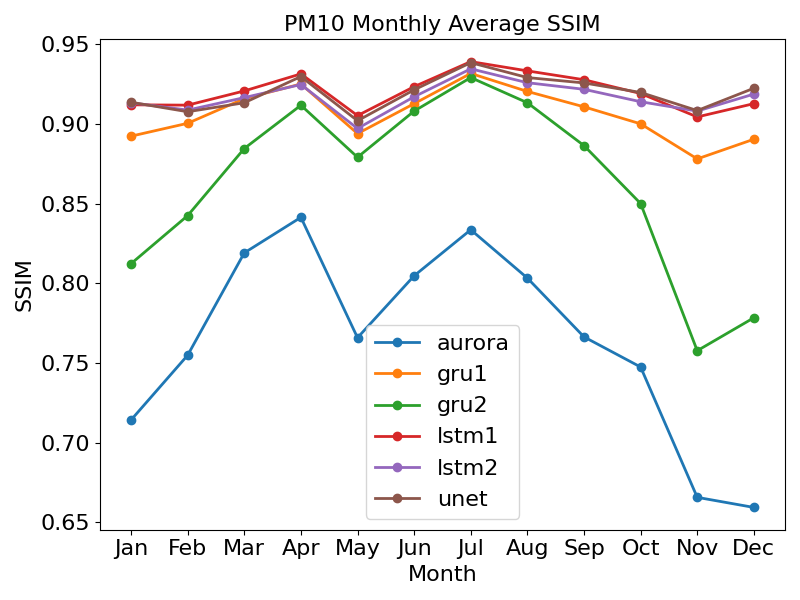}
        \caption{PM$_{10}$ SSIM}
    \end{subfigure}
    \caption{Monthly averages of normalized RMSE (NRMSE = RMSE/mean) and SSIM for PM$_{1}$, PM$_{2.5}$, and PM$_{10}$ across all models. Each row shows a PM type with NRMSE on the left and SSIM on the right.}
    \label{fig:pm_metrics_sidebyside}
\end{figure}

Figure~\ref{fig:pm_metrics_sidebyside} shows monthly normalized RMSE (NRMSE = RMSE/mean) and SSIM trends for all models across PM$_1$, PM$_{2.5}$, and PM$_{10}$. NRMSE exhibits clear seasonality, with errors peaking between March and June, coinciding with the pre-monsoon period when synoptic variability, dust intrusions, and stronger thermal gradients increase particulate heterogeneity. Errors decline during the monsoon and remain comparatively stable through winter. SSIM remains consistently high across all parameters, with only minor dips during the high-variability months. Among the models, ConvLSTM$_{2}$ and U-Net demonstrate the most stable performance, maintaining low NRMSE fluctuations and high SSIM, indicating strong structural preservation and robust generalization across atmospheric regimes.

\subsection{Analysis of Spatial Bias Patterns}

\begin{figure}[hp!]
    \centering
    \begin{subfigure}{0.32\textwidth}
        \includegraphics[width=\textwidth]{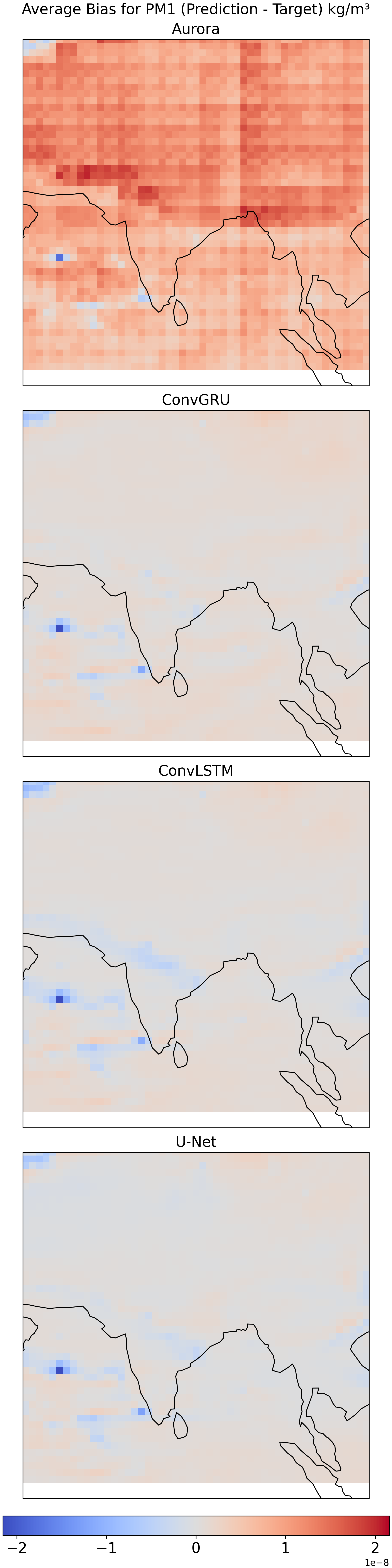}
        \caption{PM$_{1}$ Average Bias}
    \end{subfigure}
    \hfill
    \begin{subfigure}{0.32\textwidth}
        \includegraphics[width=\textwidth]{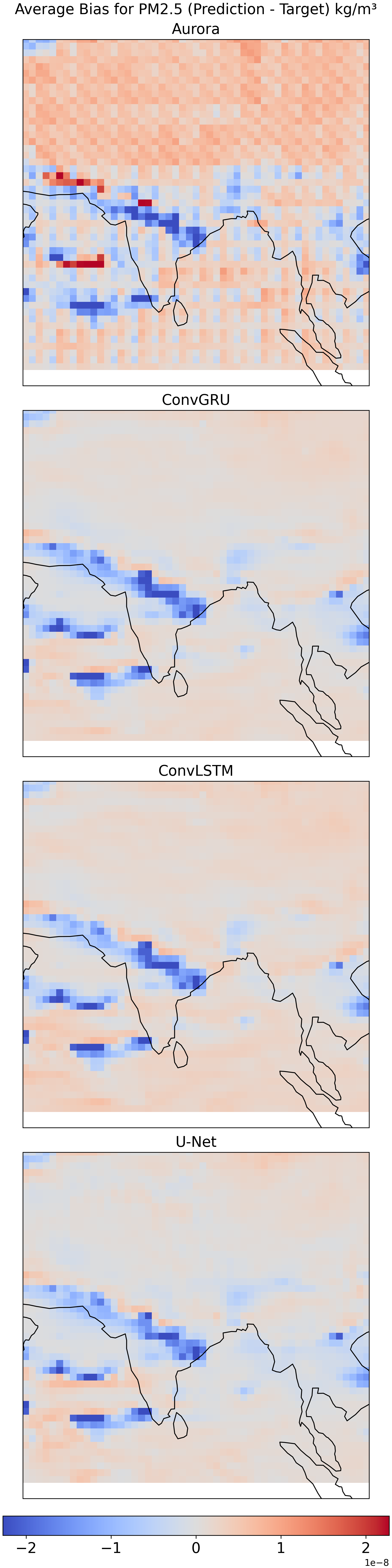}
        \caption{PM$_{2.5}$ Average Bias}
    \end{subfigure}
    \hfill
    \begin{subfigure}{0.32\textwidth}
        \includegraphics[width=\textwidth]{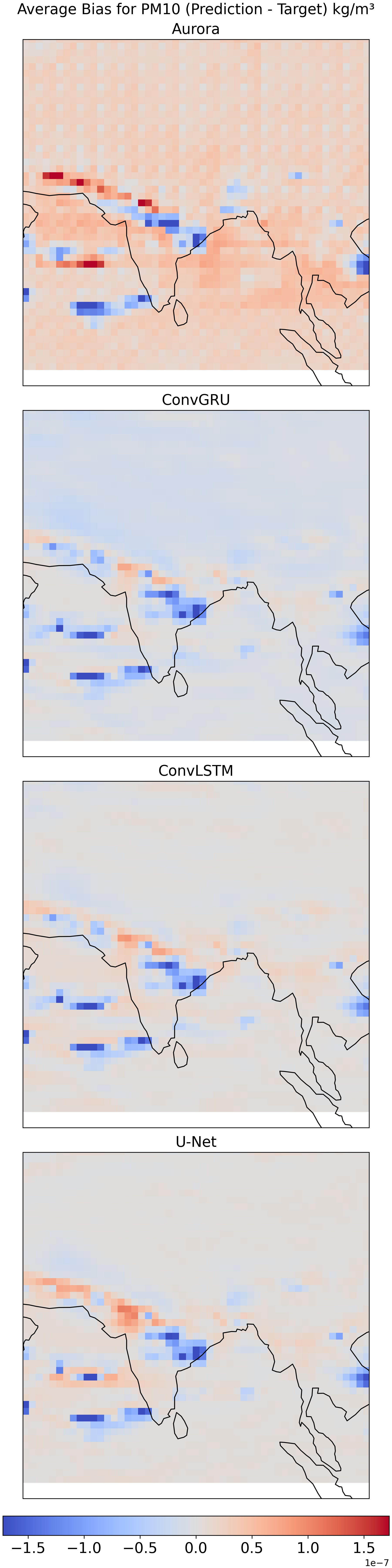}
        \caption{PM$_{10}$ Average Bias}
    \end{subfigure}
    \caption{Spatial mean bias (Prediction $-$ Target) for PM$_{1}$, PM$_{2.5}$, and PM$_{10}$, averaged over the 2024 test period. Positive (red) values indicate overestimation and negative (blue) values indicate underestimation.}
    \label{fig:pm_bias}
\end{figure}

Figure~\ref{fig:pm_bias} shows the spatially averaged bias for PM$_1$, PM$_{2.5}$, and PM$_{10}$. Aurora exhibits strong positive bias across the Indo-Gangetic Plain and northern India, indicating consistent overestimation in regions with complex aerosol dynamics. In contrast, all proposed models—ConvGRU, ConvLSTM, and U-Net—substantially reduce both the magnitude and spatial extent of bias. ConvGRU generally suppresses broad regional overprediction, while ConvLSTM more effectively corrects localized hotspots, especially for PM$_{10}$. U-Net achieves the most uniform bias reduction, maintaining near-zero deviations across the domain.

Overall, the proposed architectures achieve notable improvements over Aurora, including 20--45\% reductions in RMSE, up to 0.35 increases in SSIM, and substantial suppression of spatial bias. Despite being three orders of magnitude smaller, the lightweight models deliver more accurate, stable, and spatially coherent six-hour particulate-matter forecasts, underscoring their suitability for efficient regional nowcasting.

\section{Conclusion}

This study introduced an efficient framework for 6-hour particulate-matter nowcasting using CAMS Global Analysis data, demonstrating that compact, parameter-specific models can outperform a large-scale foundation model in regional forecasting tasks. By leveraging a fixed 256$\times$256 spatial context and predicting over a 128$\times$128 domain centered on the Indian subcontinent, the proposed approach effectively captures regional aerosol dynamics despite coarse-resolution global inputs. Across particulate matter types—PM$_1$, PM$_{2.5}$, and PM$_{10}$—the ConvGRU, ConvLSTM, and U-Net architectures consistently achieved lower errors, reduced bias, and higher structural similarity than the Aurora model, while requiring three orders of magnitude fewer parameters. These gains highlight the advantages of localized learning, model specialization, and computational efficiency for short-term air-quality prediction. The results confirm that targeted, lightweight architectures provide a practical and accurate alternative to large foundation models for regional atmospheric forecasting, offering fast and reliable operational applicability.

\section*{Limitations and Future Work}

This work is limited by the exclusive use of surface-level variables and coarse 0.4$^{\circ}$ CAMS analyses, which may underrepresent vertical aerosol dynamics and local emission gradients. Future work will incorporate additional meteorological and chemical variables, explore multi-time-step inputs, and extend the approach to higher-resolution regional datasets. Lightweight multi-output or transformer-based architectures also present promising directions for enhanced scalability and accuracy.

\bibliography{sample}

\end{document}